\documentclass{article}
\usepackage{ijcai26}

\usepackage{times}
\usepackage{soul}
\usepackage{url}
\usepackage[hidelinks]{hyperref}
\usepackage[utf8]{inputenc}
\usepackage[small]{caption}
\usepackage{graphicx}
\usepackage{subcaption}
\usepackage{amsmath}
\usepackage{amsthm}
\usepackage{booktabs}
\usepackage{algorithm}
\usepackage{algorithmic}
\usepackage[switch]{lineno}

\usepackage{tikz}
\usetikzlibrary{positioning, arrows.meta, calc, patterns}

\usepackage{multirow}  
\usepackage[table,xcdraw]{xcolor}  

\title{TTKV: Temporal-Tiered KV Cache for Long-Context LLM Inference}

\begin{document}
\author{Gradwell Dzikanyanga$^{1}$†, Weihao Yang$^{1}$, Hao Huang$^{1}$, Donglei Wu$^{2}$, Shihao Wang$^{1}$, \\
Wen Xia$^{1}$‡\thanks{Corresponding Author Wen Xia.}, Sanjeeb K C$^{1}$ \\
$^{1}$ Harbin Institute of Technology, Shenzhen, $^{2}$ Guangzhou University \\
†: \texttt{gdzikanyanga@gmail.com}, ‡: \texttt{xiawen@hit.edu.cn}}

\maketitle


\begin{abstract}
    Key–value (KV) caching is critical for efficient inference in large language models (LLMs), yet its memory footprint scales linearly with context length, resulting in a severe scalability bottleneck.
    Existing approaches largely treat KV states as equally important across time, implicitly assuming uniform precision and accessibility. However, this assumption contrasts with human memory systems, where memories vary in clarity, recall frequency, and relevance with temporal proximity.
    Motivated by this insight, we propose \textbf{TTKV}, a KV cache management framework that maps the human memory system onto the KV cache. TTKV partitions the KV cache into temporal tiers with heterogeneous capacity and precision. The design addresses three aspects: (1) \textbf{Tier Layout}, decoupling fast and slow memory using HBM and DRAM; (2) \textbf{Tier Content}, assigning more recent KV states to faster, higher-precision tiers based on temporal proximity; and (3) \textbf{Tier Interaction}, employing block-wise streaming attention to overlap communication and computation when accessing slow tiers.
    Experiments show that TTKV reduces cross-tier traffic by $5.94\times$ on 128K-context tasks, achieving up to 76\% latency reduction and $2\times$ throughput improvement over strong baselines.

\end{abstract}

\section{Introduction}

Large language models (LLMs) have demonstrated remarkable capabilities across a wide range of tasks~\cite{xiao2024efficientstreaminglanguagemodels}. A key mechanism enabling efficient autoregressive inference is the KV cache, which stores the key and value activations of previously generated tokens to avoid recomputation in transformer attention layers\cite{hooper2025kvquant10millioncontext,liu2024kivi}.


However, the KV cache grows linearly with context length, becoming a dominant bottleneck in both memory consumption and inference latency. For example, with a 32K token window on LLaMA-7B, the KV cache alone can require tens of gigabytes of memory~\cite{wang2025fierfinegrainedefficientkv}.
In standard LLM inference pipelines, inference consists of two stages: a parallel \emph{prefill} stage that processes the input and populates the KV cache, and a sequential \emph{decode} stage that generates tokens autoregressively by repeatedly querying cached KVs~\cite{pmlr-v202-sheng23a,10.1145/3651890.3672274}. During decoding, each step attends over the entire cached context, making KV cache access a primary determinant of per-token latency.


The KV cache bottleneck becomes more significant once its size exceeds high-bandwidth memory (HBM) and should be spilled into host memory (DRAM)~\cite{pmlr-v202-sheng23a,10.1145/3651890.3672274,sharma-etal-2025-minikv}. Since PCIe bandwidth is typically an order of magnitude lower than HBM bandwidth, cross-tier transfers dominate latency in long-context decoding~\cite{jiang-etal-2025-kvpr}. Consequently, the performance bottleneck shifts from merely reducing KV cache size to effectively integrating KV data-reduction techniques, such as quantization and sparsity, with the memory hierarchy to minimize costly cross-tier reads. Without such integration, these techniques remain constrained by host-to-GPU traffic when KV caches are offloaded to slower memory.


Existing approaches to mitigate this bottleneck fall into two categories: (1) \textbf{KV reduction}, which reduces the KV cache footprint through quantization and sparsification~\cite{hooper2025kvquant10millioncontext,liu2024kivi,zhang2023h2oheavyhitteroracleefficient}; and (2) \textbf{KV offloading}, which offloads the KV cache to DRAM and optimizes access patterns to reduce transfer costs~\cite{pmlr-v202-sheng23a,sharma-etal-2025-minikv}.
However, neither approach alone fully resolves the challenges of long-context LLM inference. KV reduction alone may still result in KV caches exceeding GPU memory, while KV offloading alone does not address the growth of the KV cache. Moreover, naively combining these techniques is non-trivial, as it often leads to conflicting trade-offs among cache size reduction, transfer volume, and decode latency. This raises the question: \emph{how to effectively combine KV reduction and KV offloading techniques to support efficient long-context LLM inference?}



To address these challenges, we propose \textbf{TTKV}, a new perspective on KV cache management inspired by human memory mechanisms. Existing approaches largely treat KV states as equally important across time, implicitly assuming uniform precision and accessibility. In contrast, our key insight is that the importance of KV states varies over time: \textbf{recent states function as short-term memory and are more critical for generation, whereas older states form long-term memory and only a small subset remains relevant to the current query.}
This abstraction enables KV caches with different temporal relevance to be managed with heterogeneous latency and precision requirements, laying the foundation for efficient long-context LLM inference.

We implement TTKV, the Temporal-Tiered KV Cache, which abstracts KV cache management through human memory mechanisms. The design consists of three key components. (1) \textbf{Tier Layout}: KV caches are organized into a fast tier (short-term memory) and a slow tier (long-term memory) aligned with modern hardware hierarchies, with latency-sensitive states placed in HBM and capacity-oriented states in DRAM. (2) \textbf{Tier Content}: KV states are allocated across tiers according to temporal relevance: the fast tier preserves full precision for recent and frequently accessed tokens, while the slow tier applies differential quantization and sparsification to older, less frequently accessed states. (3) \textbf{Tier Interaction}: To mitigate the latency overhead introduced by the slow tier, we employ streaming attention that overlaps computation and communication, enabling efficient access to DRAM-resident KV states during decoding.

We evaluate TTKV on a diverse set of LLMs, including LLaMA-3.1-8B~\cite{grattafiori2024llama3herdmodels}, Qwen2.5-32B~\cite{hui2024qwen25codertechnicalreport}, DeepSeek-R1-14B~\cite{zheng-etal-2025-processbench}, and LLaMA-3.1-70B~\cite{grattafiori2024llama3herdmodels}. We compare TTKV with state-of-the-art KV cache optimization methods, including KIVI~\cite{liu2024kivi}, KVQuant~\cite{hooper2025kvquant10millioncontext}, DiffKV~\cite{zhang2025diffkvdifferentiatedmemorymanagement}, and ShadowKV~\cite{sun2025shadowkvkvcacheshadows}, in long-context inference settings.
Across these models, TTKV consistently outperforms prior approaches. In particular, on 128K-context tasks, TTKV reduces cross-tier traffic by up to 5.94×, leading to as much as 76\% latency reduction and 2× throughput improvement over compared methods, while maintaining model accuracy.


In a nutshell, we make the following contributions:
\begin{itemize}
    \item We observe that existing KV cache optimization methods typically emphasize either KV reduction or KV offloading, yet often fail to effectively mitigate cross-tier traffic and end-to-end decode latency in long-context inference while preserving model accuracy.
    
    \item We propose \emph{TTKV}, a new perspective on KV cache management inspired by human memory mechanisms. TTKV partitions the KV cache into temporal tiers with heterogeneous capacity and precision, and is realized through tier layout, tier content, and tier interaction.
    
    \item we demonstrate that \emph{TTKV} significantly reduces host-to-GPU traffic and end-to-end decode latency across a wide range of context lengths, while preserving model accuracy and enabling scalable long-context LLM inference compared to state-of-the-art methods.

\end{itemize}

\section{Related Work}


\paragraph{KV Reduction Methods} aims to mitigate GPU memory overhead through quantization or sparsification. Specifically, quantization compresses KV states into low-precision representations to shrink the cache footprint~\cite{liu2024kivi,hooper2025kvquant10millioncontext,jiang2026packkvreducingkvcache}. Sparsification limits memory I/Os by selectively retaining or retrieving only salient KV states during attention computation~\cite{kitaev2020reformerefficienttransformer,xiao2024efficientstreaminglanguagemodels,liu2026freekvboostingkvcache}.
However, under long-context workloads, the KV cache size still scales linearly with the context length and exceeds the GPU memory.


\textbf{KV Offloading Methods} aim to extend KV cache capacity by offloading KV states from GPU memory to slower but larger host DRAM~\cite{pmlr-v202-sheng23a,sun2025shadowkvkvcacheshadows,jiang-etal-2025-kvpr}. These approaches primarily optimize memory placement strategies and data transfer scheduling methods to alleviate GPU memory pressure. However, without joint optimizations with KV reduction techniques, they still incur substantial cross-tier traffic and fail to minimize end-to-end decode latency in long-context inference.

\section{Design Principle}
\label{sec:motivation}

Human memory mechanisms often assign different levels of importance to memories over the time dimension. Therefore, we mimic human memory mechanisms by explicitly decoupling the model's KV cache management into two-tier implementations.
Based on the above ideas, we propose a key structure called \textbf{Tier}, considering following three aspects:

\begin{enumerate}
    \item \textbf{Tier Layout:} Tiers will be placed in accordance with current computer architectures (e.g., HBM or DRAM). 
    \item \textbf{Tier Content:} Different tiers will have different KV cache capacities and precision (bits) to to simulate the differentiated importance over time dimension.
    \item \textbf{Tier Interaction:} Two tiers will interact efficiently through a pipeline approach at both prefilling and decoding, with overlapped computation and communication.
\end{enumerate}

We now present empirical evidence validating the three design perspectives of \textbf{Tier} and demonstrating their necessity for efficient and scalable long-context LLM inference.


\begin{figure}[t]
  \centering
  \captionsetup{skip=0pt}
  \includegraphics[width=\linewidth]{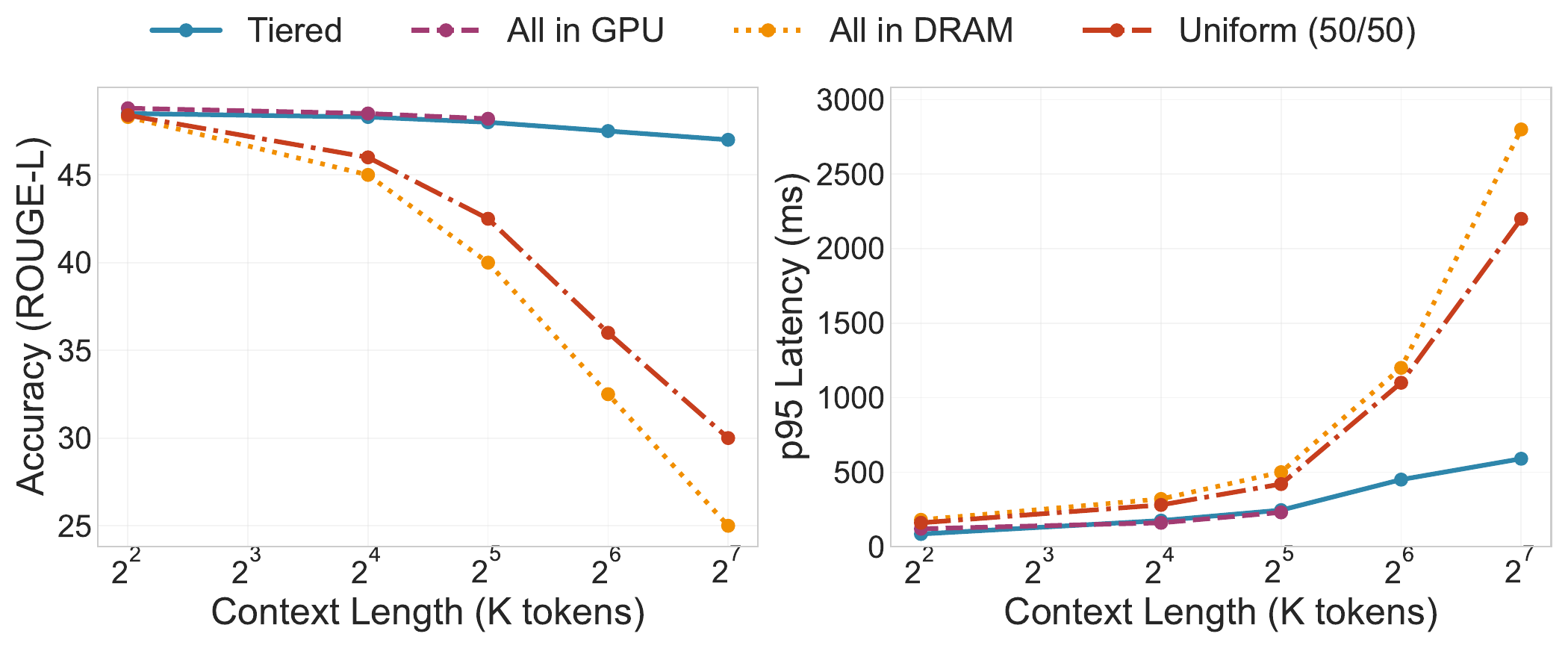}
    \caption{Accuracy and latency trade-offs for different KV cache placement strategies. Only the tiered layout (i.e., TTKV) maintains both high accuracy and low latency as context length scales.}
    \label{fig:cascadekv_ablation}
\end{figure}

\paragraph{Tier Layout: Placement According to Architecture.} 
Tier layout can lead to longer context length and better efficiency.
We validate this with four placement strategies for the KV cache in Llama-3.1-8B: storing all entries in GPU HBM, all in host DRAM, a naive uniform split, and our proposed tiered placement. As shown in Figure~\ref{fig:cascadekv_ablation}, the GPU-only strategy fails at long contexts due to memory capacity limits, while the DRAM-only approach preserves capacity but suffers from prohibitive latency. The uniform partition performs poorly on both metrics. However, only the \textbf{tiered layout---which retains recent, frequently accessed tokens in HBM while offloading older contexts to DRAM}---simultaneously maintains model accuracy and low latency. This confirms that tier placement aligned with computer memory hierarchy is essential for scalable long-context inference.

\paragraph{Tier Content: Differential Capacity and Precision.} 
Human memory prioritizes critical information with high fidelity while compressing less essential details. This principle directly informs our tiered precision strategy: the tier for fast memory (i.e., HBM) retains high precision for recent, frequently accessed tokens—analogous to vivid, immediate recollections—while the tier for slow memory (i.e., DRAM) employs aggressive compression for older, less-accessed contexts, mirroring how the brain archives distant memories.

\begin{figure}[t]
  \centering
  \captionsetup{skip=0pt}
  \includegraphics[width=\linewidth]{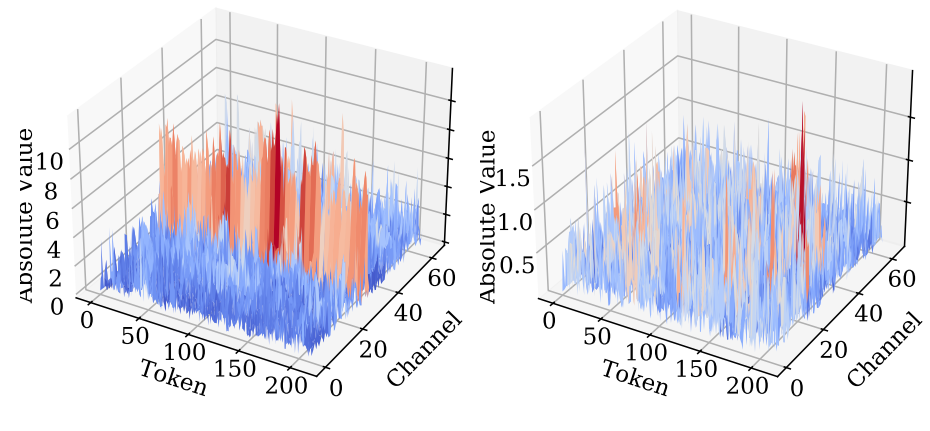}
    \caption{Magnitude distributions of keys (left) and values (right) in Llama-3.1-8B. Keys show greater dynamic range, justifying higher precision allocation in the fast-tier compared to the slow-tier.}
    \label{fig:key_value_magnitude}
\end{figure}

To further improve the compression ratio of the slow tier without sacrificing attention accuracy, we examine the inherent sensitivity of keys and values. Figure~\ref{fig:key_value_magnitude} reveals that in Llama-3.1-8B, keys exhibit significantly greater variance and dynamic range than values, making them more susceptible to precision loss. As such, within the compressed slow-tier, we allocate more bits to accuracy-critical keys and fewer to more tolerant values, thereby maximizing storage efficiency while preserving the fidelity of attention computations.

\begin{figure}[t]
  \centering
  \captionsetup{skip=0pt}
  \includegraphics[width=\linewidth]{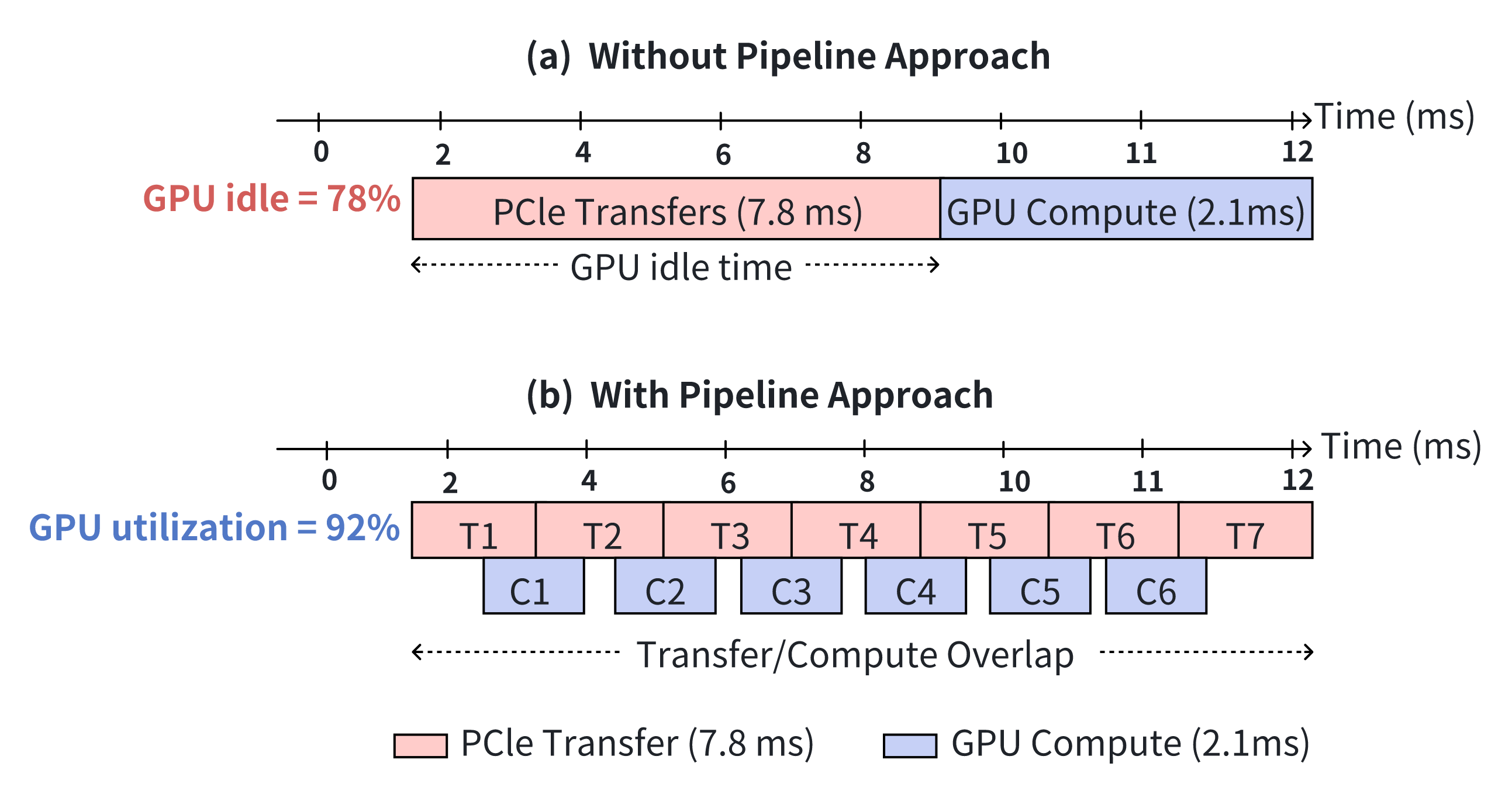}
    \caption{Timeline of a decoding step (64K context, batch=8). (a) Baseline: bulk transfers with GPU idle. (b) Pipeline: communication and computation are overlapped for better GPU utilization.}
    \label{fig:timeline_analysis}
\end{figure}

\paragraph{Tier Interaction: Pipelining Computation and Communication}.  
Human memory retrieval does not halt cognitive processing; instead, it streams relevant details incrementally while reasoning continues. This efficient recall mechanism motivates a pipeline transmission for cross-tier interaction, where communication overlaps with computation. We profile a 64K-context decoding step in Llama-3.1-8B (Figure~\ref{fig:timeline_analysis}). The baseline approach performs bulk transfers that serialize computation and communication, leaving the GPU idle for approximately 78\% of the decode time. 
By simply implementing a pipeline that breaks transfers into smaller chunks and overlaps them with attention computation, we reduce effective transfer latency by approximately 3$\times$. This demonstrates that overlapping computation and communication is essential for minimizing the performance impact of cross-tier data movement, indicating that a pipelined attention method for inter-tier interaction can ensure efficiency.

\begin{figure}[t]
  \centering
  \captionsetup{skip=0pt}
  \includegraphics[width=\linewidth]{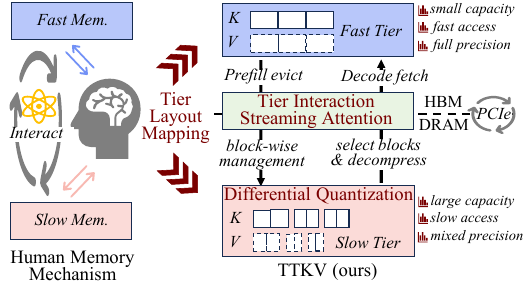}
    \caption{\textbf{The overview of TTKV.} 
    Inspired by human memory mechanism, TTKV consists of three parts: \textit{Tier Layout Mapping}, \textit{Tier Content} (\textit{Differential Quantization}), and \textit{Tier Interaction} (\textit{Streaming Attention}), which enable efficient inference.
    }
    \label{fig:overview}
\end{figure}

\section{Methodology}
\label{methodology}

Following the three design principles established in Section~\ref{sec:motivation}—Tier Layout, Tier Content, and Tier Interaction—we present \textbf{TTKV}, a unified framework that co‑designs KV cache management with the memory hierarchy. As shown in Figure~\ref{fig:overview}, TTKV explicitly implements: (1) a two‑tier memory layout (Section~\ref{sec:tiered_memory}), (2) differential quantization (Section~\ref{sec:content_of_tiers}), and (3) streaming attention (Section~\ref{sec:interaction_between_tiers}).

\subsection{Tier Layout: Tiered Memory Architecture}
\label{sec:tiered_memory}

The \textbf{Tier Layout} principle aligns the memory placement strategy with computer architecture: the \textbf{fast-tier} resides in GPU HBM for low-latency access and stores KVs in full precision, while the \textbf{slow-tier} uses host DRAM for higher capacity and stores KVs in compressed form. As discussed in Section~\ref{sec:motivation}, this two-tier partitioning ensures that recent, frequently accessed tokens remain in fast memory, while older context is stored in slower memory without sacrificing model accuracy. TTKV's layout is shown in Figure~\ref{fig:overview}. 

\begin{itemize}
    \item \textbf{Fast-tier (GPU HBM):} Rather than fixing the block count for the fast-tier, we dynamically define its contents based on the available GPU HBM capacity. Specifically, let \(L_{\text{fast}}\) denote the number of most recent tokens stored in full precision (FP16) in GPU HBM. The size \(L_{\text{fast}}\) is chosen such that the memory footprint \(M_{\text{fast}}\) does not exceed the allocated HBM capacity \(C_{\text{HBM}}\):
    \[
    M_{\text{fast}} = L_{\text{fast}} \times d_{\text{kv}} \times b_{\text{FP16}} \leq C_{\text{HBM}},
    \]
    where \(d_{\text{kv}}\) is the per-token KV feature dimension, and \(b_{\text{FP16}}\) is the byte width of FP16 precision. Therefore, the fast-tier (i.e., HBM) holds the most recent \(L_{\text{fast}}\) tokens in full precision, ensuring that the most frequently accessed data are cached in the fastest memory.

     \item \textbf{Slow-tier (Host DRAM):} The slow-tier stores older tokens in a compressed form, with a larger memory capacity but slower access speed. Moreover, these tokens are partitioned into structured blocks for efficient memory management and cross-tier movement.
\end{itemize}

Formally, let \(\mathcal{K}\) and \(\mathcal{V}\) denote the full KV cache for a given context. TTKV partitions the cache as:
\[
\mathcal{K} = \mathcal{K}_{\text{fast}} \cup \mathcal{K}_{\text{slow}}, \quad 
\mathcal{V} = \mathcal{V}_{\text{fast}} \cup \mathcal{V}_{\text{slow}},
\]
where the subscripts \(\text{fast}\) and \(\text{slow}\) denote entries assigned to the fast and slow tiers, respectively. Entries in \(\mathcal{K}_{\text{fast}}\) and \(\mathcal{V}_{\text{fast}}\) are selected based on recency and role importance (e.g., high attention scores), whereas less frequently accessed or lower-importance entries are placed in \(\mathcal{K}_{\text{slow}}\) and \(\mathcal{V}_{\text{slow}}\). This selection maximizes the fraction of accesses served from GPU HBM, thereby reducing costly PCIe transfers and preserving accuracy for the most impactful tokens.

\textbf{Block-wise Organization in Slow-tier.}  
The block structure is a core mechanism enabling the \textbf{Tier Interaction}. By organizing the slow‑tier KV cache into fixed‑size \emph{blocks} of $B_{\text{blk}}$ tokens, we create manageable units for the streaming attention, allowing fine‑grained scheduling that overlaps transfer and computation. This mirrors the cognitive process of \textit{chunking}, where the brain groups information into units for efficient recall and processing.
We experimented with block sizes \(B_{\text{blk}} \in \{32, 64, 128, 256\}\) and found that 128 offered the best trade-off between latency and accuracy across models; details are in Section~\ref{subsec:ablation}.

Each slow‑tier block $B_j$ contains compressed representations of $B_{\text{blk}}$ key and value vectors. Moreover, a lightweight index, functioning like a cognitive pointer to memory chunks, maps token positions to block identifiers:
\begin{equation}
  \mathrm{idx}: \mathcal{T}_\text{slow}(t) \to \{1,2,\ldots\},
  \label{eq:block-index}
\end{equation}
so that, given a position $i \in \mathcal{T}_\text{slow}(t)$, $\mathrm{idx}(i)$ returns the specific block $B_j$. This index structure is fundamental to the \textbf{Tier Interaction}, as it (i) provides a deterministic, fine‑grained unit for cross‑tier data fetches, and (ii) enables the \textbf{streaming attention} to efficiently locate, schedule, and transfer only the required blocks, thereby minimizing overhead and supporting the overlap of computation and communication.

\textbf{Dynamic Tier Management.}  
The flow of data between tiers is governed by the principles of \textbf{Layout} and \textbf{Content}. Newly generated tokens—representing the most immediate "working memory"—are first appended to the fast-tier in full precision. When the fast-tier reaches its capacity $C_\text{fast}$, the oldest block of $B_{\text{blk}}$ tokens is evicted. This block is then compressed according to the differential  quantization scheme (see Section \ref{sec:content_of_tiers}) before being appended to the slow-tier.

This First-In-First-Out (FIFO) management policy ensures that the working set of the attention operation remains in GPU HBM, while older contexts are preserved in a compressed, structured form in host DRAM. Moreover, this process mimics the cognitive mechanism of \textit{human memory consolidation}, in which recent experiences are initially vivid, and later are processed (e.g., compressed) for long-term storage.

\subsection{Tier Content: Differential Quantization}
\label{sec:content_of_tiers}
The \textbf{Tier Content} principle dictates that different data should be stored with different precision, mirroring the selective retention characteristics of human memory. Moreover, our analysis in Figure~\ref{fig:key_value_magnitude} suggests that \textbf{keys} exhibit larger variance and govern attention scores, whereas \textbf{values} are stable. Applying uniform compression is therefore suboptimal, which wastes bits on tolerant values and degrades critical keys.

TTKV implements this principle through \textbf{differential  quantization}, aligning precision with functional role:
\begin{itemize}
    \item \textbf{Keys} are preserved with higher precision (8-bit) to maintain the accuracy of attention score computation, which is the core component of the model's reasoning.
    \item \textbf{Values} are Compressed to lower precision (4-bit) to maximize reductions in storage footprint and cross‑tier transfer volume, with minimal impact on output quality.
\end{itemize}

Formally, for a vector \(x\) in the slow-tier, the applied differential quantization operator \(Q\) is defined as:
\[
Q(x) = \begin{cases}
Q_{8\text{-bit}}(x) & \text{if } x \in \mathcal{K}_\text{slow},\\[4pt]
Q_{4\text{-bit}}(x) & \text{if } x \in \mathcal{V}_\text{slow},
\end{cases}
\]
where \(Q_{8\text{-bit}}\) and \(Q_{4\text{-bit}}\) are quantizers for their respective bit-widths. This targeted precision assignment is a pivotal co-design element: by reducing the data volume of values before they are placed in the slow-tier, differential quantization directly amplifies the efficiency gains of the \textbf{Layout} (smaller blocks to move) and \textbf{Interaction} (less traffic to schedule) principles, while preserving model accuracy.

\subsection{Tier Interaction: Streaming Attention}
\label{sec:interaction_between_tiers}

The \textbf{Tier Interaction} principle is realized through \textbf{streaming attention} that shifts cross‑tier data movement from a blocking cost to a hidden overhead. As Figure~\ref{fig:timeline_analysis} quantifies, naive access patterns leave the GPU idle during PCIe transfers. Inspired by the brain's ability to stream and process memories concurrently, TTKV's pipeline proactively manages data flow to overlap communication with computation.

The core mechanism, detailed in Algorithm~\ref{alg:ttkv_pipeline}, orchestrates the fetch and execution of compressed KV blocks in three phases that operate in a pipelined manner:
\begin{enumerate}
    \item \textbf{Block Identification:} Based on the current query and the block index, streaming attention identifies a subset of the most relevant compressed blocks in the slow‑tier (i.e., streaming attention is a sparse attention manner).
    \item \textbf{Asynchronous Prefetch:} It issues non‑blocking requests to transfer these compressed blocks from host DRAM to GPU HBM \textit{ahead} of their immediate need.
    \item \textbf{Scheduled Overlap:} These transfers are processed in parallel with the attention computation on data already present in the fast‑tier, thus hiding the PCIe latency.
\end{enumerate}

The \textbf{Interaction} pipeline is essential to unlock the full potential of the \textbf{Layout} and \textbf{Content} principles. The tiered layout provides the structure for block‑wise management, while differential quantization reduces the volume of data to be moved. In turn, the pipeline efficiently hides the remaining transfer cost, ensuring that the theoretical benefits of the first two principles translate into real‑world latency reduction.

Formally, during each decoding step \(t\), the system state is defined by the tokens currently in the fast-tier, \(\mathcal{K}_\text{fast}^{(t)}\) and \(\mathcal{V}_\text{fast}^{(t)}\). TTKV's streaming attention executes the attention computation concurrently with data prefetching:
\[
\begin{aligned}
\mathbf{o}_t &= \mathrm{Attn}\big(Q_t,\, \mathcal{K}_\text{fast}^{(t)},\, \mathcal{V}_\text{fast}^{(t)}\big) \\
&\text{ while asynchronously fetching } \mathcal{B}_\text{next} \subset \{\mathcal{K}_\text{slow}, \mathcal{V}_\text{slow}\}.
\end{aligned}
\]
Here, \(\mathrm{Attn}(\cdot)\) computes the attention output using available data, and \(\text{AsyncPrefetch}(\cdot)\) proactively fetches the next block \(\mathcal{B}_\text{next}\) from the slow-tier based on the block index \(\mathrm{idx}\).

\begin{algorithm}[tb]
    \caption{TTKV Streaming Attention}
    \label{alg:ttkv_pipeline} 
    \textbf{Input}: Query $\mathbf{q}_t$, fast-tier cache $(\mathcal{K}_\text{fast}, \mathcal{V}_\text{fast})$, slow-tier blocks $\{B_j\}$ \\
    \textbf{Parameter}: Capacity $L_{\text{fast}}$, scoring function $\phi(\cdot)$ \\
    \textbf{Output}: Attention output $\mathbf{o}_t$
    \begin{algorithmic}[1]
        \STATE $\mathbf{o}_t \gets \mathrm{Attn}(\mathbf{q}_t, \mathcal{K}_\text{fast}, \mathcal{V}_\text{fast})$ \hfill \COMMENT{\textbf{Layout}}
        \STATE $J_{\text{scores}} \gets [\,]$
        \FOR{each block $B_j$ in slow-tier}
            \STATE append $(\phi(\mathbf{q}_t, B_j), j)$ to $J_{\text{scores}}$ \COMMENT{Identify via $\mathrm{idx}$}
        \ENDFOR
        \STATE $J_{\text{top}} \gets \text{TopK}(J_{\text{scores}}, k)$
        \FOR{each $(score, j)$ in $J_{\text{top}}$}
            \STATE $\text{AsyncPrefetch}(B_j)$ \hfill \COMMENT{\textbf{Interaction}}
            \STATE $(\widetilde{\mathcal{K}}_j, \widetilde{\mathcal{V}}_j) \gets \text{Decompress}(B_j)$ \COMMENT{\textbf{Content} (8b/4b)}
            \STATE $\mathbf{o}_t \gets \mathbf{o}_t + \mathrm{Attn}(\mathbf{q}_t, \widetilde{\mathcal{K}}_j, \widetilde{\mathcal{V}}_j)$
        \ENDFOR
        \IF{$\mathrm{len}(\mathcal{K}_\text{fast}) \geq L_{\text{fast}}$}  
            \STATE $B_{\text{evict}} \gets \text{GetOldestBlock}(\mathcal{K}_\text{fast}, \mathcal{V}_\text{fast})$
            \STATE $\text{CompressAndAppendToSlowTier}(B_{\text{evict}})$ \COMMENT{Evict \& compress}
            \STATE $\text{RemoveFromFastTier}(B_{\text{evict}})$
        \ENDIF
        \RETURN $\mathbf{o}_t$
    \end{algorithmic}
\end{algorithm}

This concurrent execution ensures that:
\begin{itemize}
    \item \textbf{GPU idle cycles are minimized} by eliminating blocking waits for data.
    \item \textbf{The effective latency of cross‑tier movement is hidden} behind ongoing, useful computation.
    \item \textbf{Necessary KV blocks are staged in HBM} just as they become relevant for subsequent attention steps.
\end{itemize}

This streaming attention is the essential enabler that translates the theoretical advantages of the Layout and Content principles into tangible latency reduction by overlapping communication with computation, directly targeting the data‑movement bottleneck in long‑context inference.

\section{Experiments}
\label{sec:experiments}

\subsection{Experimental Setup}

\label{sec:exp-setup}

\noindent\textbf{Models.} We evaluate TTKV on a set of large decoder-only language models spanning various scales and architectural families. Our primary models include LLaMA-3.1-8B and LLaMA-3.1-70B from the LLaMA2/3 series~\cite{touvron2023llama2openfoundation}, the Qwen2.5 family of models, up to 32B parameters, as described in the Qwen2.5-Coder technical report~\cite{hui2024qwen25codertechnicalreport}, and DeepSeek-R1-14B, distilled variants derived from reasoning-optimized backbones~\cite{zheng-etal-2025-processbench}. We also include Mistral-7B, representative of efficient open-weight LLMs~\cite{jiang2023mistral7b}.

\noindent\textbf{Baselines and Datasets.} We compare TTKV against the standard FP16 dense-prefix KV cache and several state-of-the-art KV-cache methods, including KIVI~\cite{liu2024kivi}, KVQuant~\cite{hooper2025kvquant10millioncontext}, DiffKV~\cite{zhang2025diffkvdifferentiatedmemorymanagement}, and ShadowKV~\cite{sun2025shadowkvkvcacheshadows}. For evaluation, we use Qasper and MultiNews from LongBench~\cite{bai-etal-2024-longbench}, the Loong extended multi-document QA benchmark~\cite{wang-etal-2024-leave}, and the RULER synthetic long-context benchmark~\cite{hsieh2024rulerwhatsrealcontext}.

\noindent\textbf{Implementation details.} All experiments are conducted on NVIDIA A100 and RTX 3090 GPUs, utilizing a unified PyTorch and CUDA implementation. Moreover, our implementation configurations are as follows: 1. measurements of memory usage, latency, and accuracy are taken under identical context lengths, with a batch size of 8 to ensure a fair comparison. 2. For our differential  quantization, we assign 8 bits for keys and 4 bits for values. 3. The KV cache in the slow-tier is organized into blocks, with each block containing 128 tokens. 4. Latency is measured on the GPU using CUDA events, with sufficient warm-up steps to ensure stable results, and p95 per-token latency is reported. 5. Memory usage is monitored using NVML counters, and host-to-GPU KV traffic is measured by aggregating host-to-GPU memory reads triggered by KV cache fetches during inference.

\begin{table}[t]
  \centering
  \scriptsize
  \setlength{\tabcolsep}{4pt}
  \renewcommand{\arraystretch}{1.2}
  \begin{tabular}{@{}cccccc@{}}
    \toprule
    \textbf{Model} & \textbf{Method} & \textbf{p95 Lat.} \textbf{(ms)} $\downarrow$ & \textbf{H$\rightarrow$G} \textbf{(GB)} $\downarrow$ & \textbf{Acc.} \textbf{(\%)} $\uparrow$ \\
    \midrule
    \multirow{6}{*}{\textbf{\rotatebox{90}{Mistral-7B}}} & FP16            & 340 & 3.20 & 59.2 \\
     & KIVI (2-bit)    & 325 & 2.75 & 58.8 \\
     & KVQuant (3-bit) & 332 & 2.95 & 59.0 \\
     & DiffKV          & 338 & 3.05 & 58.8 \\
     & ShadowKV        & 345 & 3.15 & 58.9 \\
     & \cellcolor{gray!15}\textbf{TTKV} & \cellcolor{gray!15}\textbf{215} & \cellcolor{gray!15}\textbf{0.95} & \cellcolor{gray!15}\textbf{59.1} \\
    \midrule
    \multirow{6}{*}{\textbf{\rotatebox{90}{Llama-3.1-8B}}} & FP16            & 380 & 3.60 & 65.4 \\
     & KIVI (2-bit)    & 362 & 3.10 & 64.9 \\
     & KVQuant (3-bit) & 370 & 3.35 & 65.2 \\
     & DiffKV          & 378 & 3.45 & 65.1 \\
     & ShadowKV        & 385 & 3.55 & 65.3 \\
     & \cellcolor{gray!15}\textbf{TTKV} & \cellcolor{gray!15}\textbf{245} & \cellcolor{gray!15}\textbf{1.05} & \cellcolor{gray!15}\textbf{65.0} \\
    \bottomrule
  \end{tabular}
  \caption{L-Eval (32K): TTKV achieves $>3\times$ lower Host-to-GPU (H$\rightarrow$G) traffic and $\sim$35\% lower p95 latency while matching FP16 accuracy.}
  \label{tab:long_context_leval_32k}
\end{table}

\begin{table}[t]
  \centering
  \scriptsize
  \setlength{\tabcolsep}{4pt}
  \renewcommand{\arraystretch}{1.2}
  \begin{tabular}{@{}lccccc@{}}
    \toprule
    \textbf{Model} & \textbf{Method} & \textbf{p95 Lat.} \textbf{(ms)} $\downarrow$ & \textbf{H$\rightarrow$G} \textbf{(GB)} $\downarrow$ & \textbf{ROUGE-L} $\uparrow$ \\
    \midrule
    \multirow{6}{*}{\rotatebox{90}{\begin{tabular}[c]{@{}c@{}}\textbf{DeepSeek-} \\ \textbf{R1-14B}\end{tabular}}} & FP16            & 285 & 2.40 & 50.2 \\
     & KIVI (2-bit)   & 275 & 2.15 & 48.8 \\
     & KVQuant (3-bit) & 280 & 2.25 & 50.0 \\
     & DiffKV & 285 & 2.30 & 49.2 \\
     & ShadowKV       & 290 & 2.38 & 49.9 \\
     & \cellcolor{gray!15} \textbf{TTKV} & \cellcolor{gray!15} \textbf{175} & \cellcolor{gray!15} \textbf{0.85} & \cellcolor{gray!15} \textbf{50.0} \\
    \midrule
    \multirow{6}{*}{\textbf{\rotatebox{90}{Llama-3.1-8B}}} & FP16            & 185 & 1.60 & 48.5 \\
     & KIVI (2-bit)   & 178 & 1.40 & 47.8 \\
     & KVQuant (3-bit) & 182 & 1.50 & 48.3 \\
     & DiffKV & 185 & 1.55 & 48.1 \\
     & ShadowKV       & 188 & 1.58 & 48.2 \\
     & \cellcolor{gray!15} \textbf{TTKV} & \cellcolor{gray!15} \textbf{118} & \cellcolor{gray!15} \textbf{0.62} & \cellcolor{gray!15} \textbf{48.4} \\
    \bottomrule
  \end{tabular}
  \caption{GovReport (16K): TTKV reduces Host-to-GPU (H$\rightarrow$G) traffic by $\sim$2.8$\times$ and p95 latency by $\sim$35\% under a 4\,GB HBM KV cache budget while preserving the ROUGE-L score.}
  \label{tab:medium_context_govreport_16k}
\end{table}



\begin{table}[t]
    \centering
    \scriptsize
    \setlength{\tabcolsep}{4pt}
    \renewcommand{\arraystretch}{1.15}
    \begin{tabular}{@{}lrrrrr@{}}
        \toprule
        \multirow{2}{*}{\textbf{Method}} & \multicolumn{5}{c}{\textbf{Context Length (tokens)}} \\
        \cmidrule(lr){2-6}
        & \textbf{4K} & \textbf{16K} & \textbf{32K} & \textbf{64K} & \textbf{128K} \\
        \midrule
        FP16                 & 135.5 & 120.4 & 89.2  & OOM  & OOM   \\
        KIVI (2‑bit)         & 142.3 & 128.0 & 95.0  & 58.7  & 29.4   \\
        KVQuant (3‑bit)      & 145.1 & 130.5 & 98.6  & 61.0  & 30.8   \\
        ShadowKV             & 150.8 & 136.7 & 105.2 & 67.5  & 34.9   \\
        \midrule
        \textbf{TTKV}   & \textbf{148.7} & \textbf{142.1} & \textbf{128.0} & \textbf{95.3} & \textbf{65.7} \\
        \bottomrule
    \end{tabular}
    \caption{Token throughput (tokens/sec) vs. context length for LLaMA‑3.1‑8B (batch size 8). Baselines show expected throughput degradation due to memory traffic, while TTKV maintains higher throughput by minimizing cross‑tier overhead.}
    \label{tab:throughput_vs_context}
\end{table}


\begin{table*}[t]
  \centering
  \scriptsize
  \setlength{\tabcolsep}{4pt}
  \renewcommand{\arraystretch}{1.15}
  \begin{tabular}{@{}llrrrrrrrr@{}}
    \toprule
    \multirow{2}{*}{\textbf{Model}} & \multirow{2}{*}{\textbf{Method}} & \multicolumn{2}{c}{\textbf{MultiNews}} & \multicolumn{2}{c}{\textbf{Qasper}} & \multicolumn{2}{c}{\textbf{RepoBench-P}} & \multicolumn{2}{c}{\textbf{RULER}} \\
    \cmidrule(lr){3-4} \cmidrule(lr){5-6} \cmidrule(lr){7-8} \cmidrule(lr){9-10}
    & & ROUGE-L $\uparrow$  & p95 (ms) $\downarrow$ & F1 score $\uparrow$ & p95 (ms) $\downarrow$ & Score $\uparrow$ & p95 (ms) $\downarrow$ & Acc.\% $\uparrow$ & p95 (ms) $\downarrow$ \\
    \midrule
    \multirow{6}{*}{\textbf{Llama-3.1-70B}} &
    FP16  & 29.5 & 2400 & 38.0 & 2450 & 58.0 & 2500 & 62.0 & 2450 \\
    & KIVI      & 27.7 & 1000 & 35.8 & 1020 & 55.8 & 1030 & 59.5 & 1015 \\
    & KVQuant   & 28.5 & 950  & 36.8 & 970  & 56.8 & 980  & 60.8 & 965  \\
    & DiffKV    & 28.7 & 850  & 37.0 & 880  & 57.0 & 890  & 61.0 & 875  \\
    & ShadowKV  & 28.5 & 820  & 36.8 & 850  & 56.8 & 860  & 60.8 & 845  \\
    & \cellcolor{gray!15}\textbf{TTKV} & \cellcolor{gray!15}\textbf{29.3} & \cellcolor{gray!15}\textbf{580} & \cellcolor{gray!15}\textbf{37.8} & \cellcolor{gray!15}\textbf{610} & \cellcolor{gray!15}\textbf{57.8} & \cellcolor{gray!15}\textbf{600} & \cellcolor{gray!15}\textbf{61.8} & \cellcolor{gray!15}\textbf{590} \\
    \midrule
    \multirow{6}{*}{\textbf{Qwen2.5-32B}} &
    FP16  & 28.0 & 2300 & 39.0 & 2350 & 60.0 & 2400 & 64.0 & 2350 \\
    & KIVI      & 26.2 & 980  & 36.5 & 1020 & 57.0 & 1050 & 60.5 & 1015 \\
    & KVQuant   & 27.2 & 920  & 37.8 & 960  & 58.2 & 980  & 62.3 & 965  \\
    & DiffKV    & 27.4 & 850  & 38.0 & 880  & 58.5 & 900  & 62.5 & 885  \\
    & ShadowKV  & 27.2 & 820  & 37.9 & 860  & 58.3 & 880  & 62.3 & 865  \\
    & \cellcolor{gray!15}\textbf{TTKV} & \cellcolor{gray!15}\textbf{27.9} & \cellcolor{gray!15}\textbf{540} & \cellcolor{gray!15}\textbf{38.9} & \cellcolor{gray!15}\textbf{570} & \cellcolor{gray!15}\textbf{59.9} & \cellcolor{gray!15}\textbf{560} & \cellcolor{gray!15}\textbf{63.9} & \cellcolor{gray!15}\textbf{550} \\
    \midrule
    \multirow{6}{*}{\textbf{DeepSeek-R1-14B}} &
    FP16  & 27.0 & 2500 & 37.5 & 2550 & 59.5 & 2600 & 63.5 & 2550 \\
    & KIVI      & 25.3 & 1020 & 35.0 & 1050 & 56.5 & 1070 & 59.8 & 1045 \\
    & KVQuant   & 26.2 & 950  & 36.2 & 980  & 58.0 & 1000 & 61.8 & 985  \\
    & DiffKV    & 26.4 & 880  & 36.5 & 910  & 58.3 & 930  & 62.1 & 915  \\
    & ShadowKV  & 26.2 & 845  & 36.3 & 880  & 58.1 & 900  & 61.9 & 885  \\
    & \cellcolor{gray!15}\textbf{TTKV} & \cellcolor{gray!15}\textbf{26.9} & \cellcolor{gray!15}\textbf{450} & \cellcolor{gray!15}\textbf{37.3} & \cellcolor{gray!15}\textbf{480} & \cellcolor{gray!15}\textbf{59.3} & \cellcolor{gray!15}\textbf{470} & \cellcolor{gray!15}\textbf{63.3} & \cellcolor{gray!15}\textbf{460} \\
    \bottomrule
  \end{tabular}
  \caption{128K long-context performance on MultiNews, Qasper, RepoBench-P, and RULER. TTKV delivers substantially lower p95 latency while maintaining near-FP16 quality.}
  \label{tab:overall-longctx-128k-final}
\end{table*}

\begin{table}[t]
    \centering
    \scriptsize
    \setlength{\tabcolsep}{3.5pt}
    \renewcommand{\arraystretch}{1.1}
    \begin{tabular}{@{}rrcc@{}}
        \toprule
        \textbf{Model} & \textbf{Method} & \textbf{H$\rightarrow$G} \textbf{(GB)} $\downarrow$ & \textbf{Avg. Lat.} \textbf{(ms)} $\downarrow$ \\
        \midrule
        \multirow{2}{*}{\textbf{Llama-3.1-70B}} 
            & FP16          & 92.0 & 2450 \\
            & \textbf{TTKV} & \textbf{15.3} & \textbf{595} \\
        \midrule
        \multirow{2}{*}{\textbf{Qwen2.5-32B}} 
            & FP16          & 92.0 & 2350 \\
            & \textbf{TTKV} & \textbf{15.5} & \textbf{555} \\
        \midrule
        \multirow{2}{*}{\textbf{DeepSeek-R1-14B}} 
            & FP16          & 92.0 & 2550 \\
            & \textbf{TTKV} & \textbf{15.8} & \textbf{465} \\
        \bottomrule
    \end{tabular}
    \caption{128K summary table: TTKV reduces H$\rightarrow$G traffic by $\sim$5.94$\times$ and p95 latency by $\sim$76\% (averaged across tasks).}
    \label{tab:summary_128k_compact_multirow}
\end{table}


\begin{figure}[t]
  \centering
  \captionsetup{skip=0pt} 
  \begin{subfigure}{0.50\columnwidth} 
    \centering
    \includegraphics[width=\linewidth]{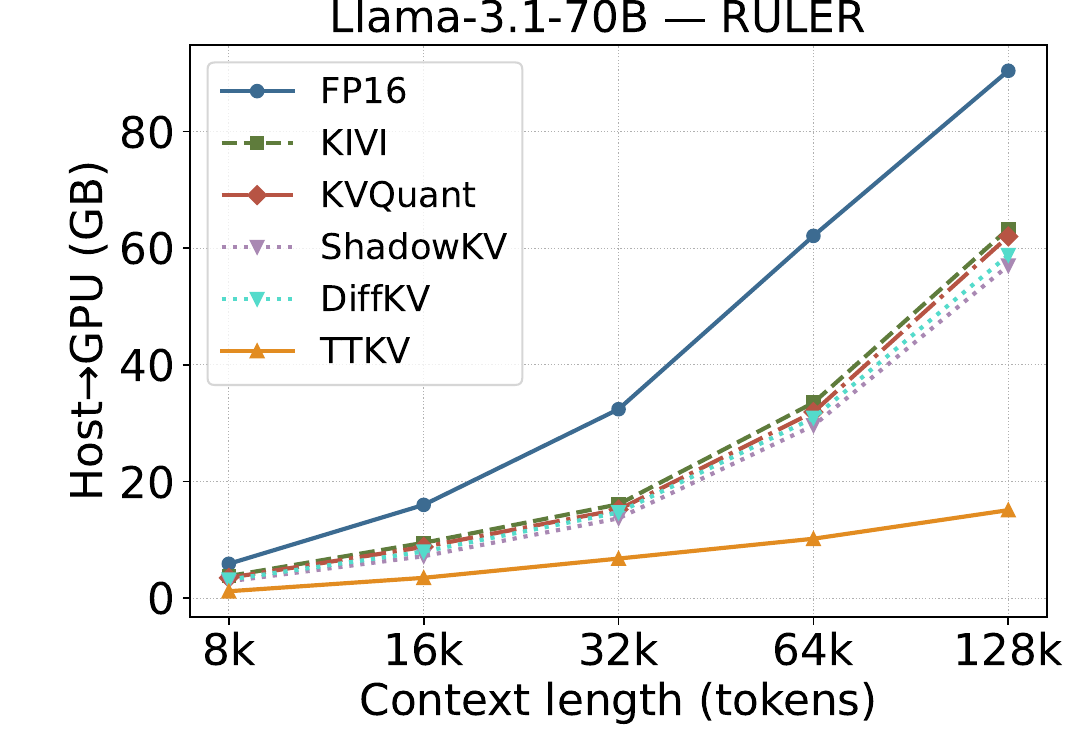}
    \caption{Host-to-GPU Traffic}
    \label{fig:kv_access_baseline}
  \end{subfigure}
  \hfill
  \begin{subfigure}{0.48\columnwidth} 
    \centering
    \includegraphics[width=\linewidth]{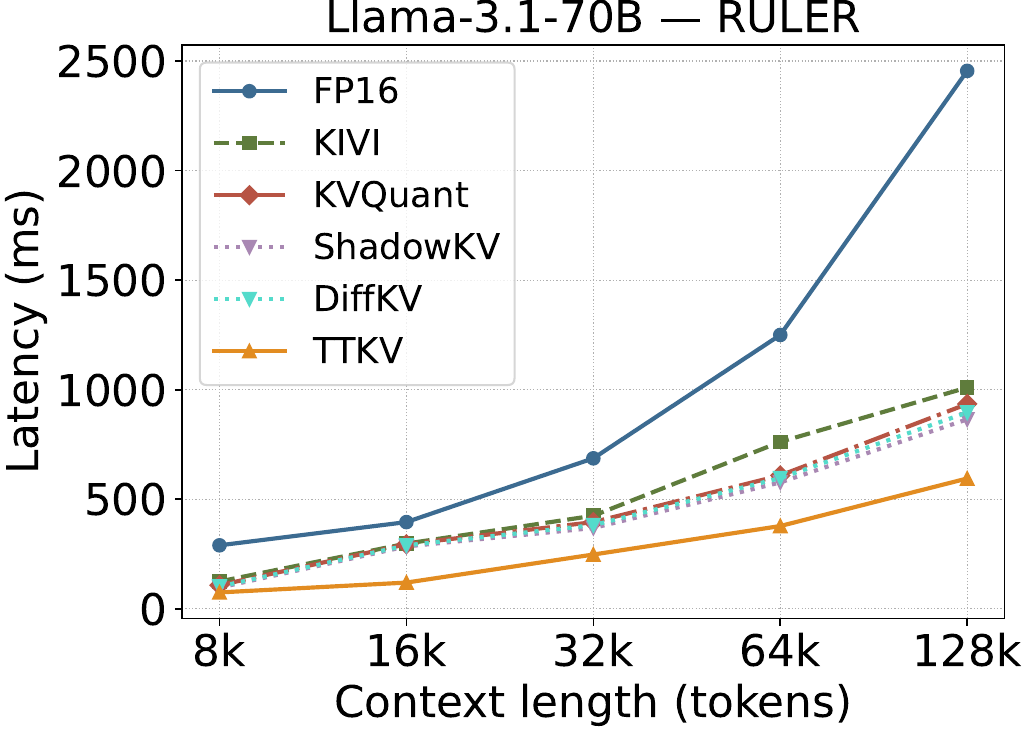}
    \caption{p95 Decode Latency}
    \label{fig:cross_tier_traffic_vs_latency}
  \end{subfigure}
  
  \caption{Performance scaling on RULER dataset.}
  \label{fig:kv_access_compare}
\end{figure}

\begin{figure}[t]
  \centering
  \captionsetup{skip=0pt}
  \includegraphics[width=\linewidth]{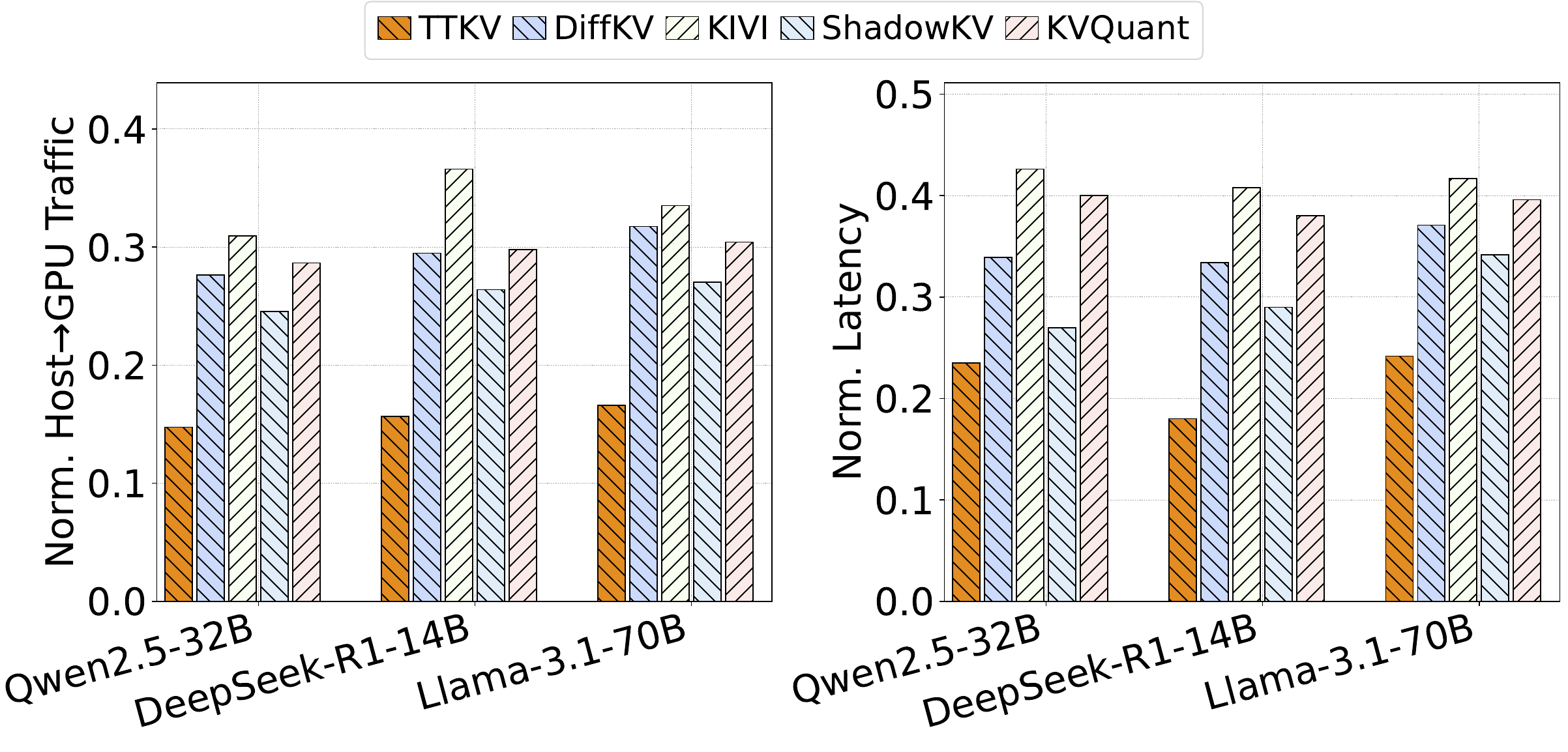}
  \caption{Normalized Host→GPU traffic and p95 latency across models. TTKV reduces both metrics compared to state-of-the-art KV compression baselines (KIVI, KVQuant, ShadowKV).}
  \label{fig:normalized_comparison}
\end{figure}

\subsection{Results and Analysis}

\paragraph{Host-to-GPU Traffic Reduction:}  
It is apparent from Figure~\ref{fig:kv_access_compare}, Figure~\ref{fig:normalized_comparison} and Table~\ref{tab:summary_128k_compact_multirow} that TTKV achieves significant reductions in host-to-GPU traffic across all evaluated models, including LLaMA-3.1-70B, Qwen2.5-32B, and DeepSeek-R1-14B. The reductions range from 2.8× to 5.94×, depending on the context length. TTKV’s efficiency is more obvious at longer context lengths, where it outperforms the baseline FP16 method by more than 5.94× in terms of host-to-GPU traffic at the 128K context. These results show that TTKV effectively manages memory traffic, especially with longer context lengths, reducing overhead in large-scale inference tasks compared to existing methods. 

\paragraph{p95 Latency Reduction:}  
Table~\ref{tab:long_context_leval_32k} and Table~\ref{tab:medium_context_govreport_16k} indicates that TTKV significantly reduces p95 latency by 35\% on average compared to FP16 across different models, including LLaMA-3.1-8B, Mistral-7B, and DeepSeek-R1-14B, evaluated on medium-context tasks such as those in the L-Eval and GovReport datasets. The largest improvements are observed in long-context tasks, particularly at the 128K context length, where TTKV reduces p95 latency by approximately 76\% compared to FP16. This is particularly evident on models like LLaMA-3.1-70B and Qwen2.5-32B, across tasks such as Qasper, MultiNews, and RULER (see Table~\ref{tab:overall-longctx-128k-final} and Figure~\ref{fig:normalized_comparison}). TTKV consistently outperforms all state-of-the-art methods, including KIVI, KVQuant, and ShadowKV, demonstrating its superior efficiency in optimizing decoding time while maintaining strong task performance.

\paragraph{Throughput Analysis}  
The data in Table~\ref{tab:throughput_vs_context} indicates that TTKV consistently achieves higher token throughput (tokens/sec) across increasing context lengths, demonstrating its superior end-to-end generation efficiency. TTKV achieves up to approximately 2× higher throughput at 128K context length compared to the best baseline, reflecting reduced cross-tier overhead and improved decoding efficiency.

\paragraph{Accuracy Preservation:}  
Results in Table~\ref{tab:long_context_leval_32k} and Table~\ref{tab:overall-longctx-128k-final} show that TTKV maintains accuracy comparable to or slightly better than state-of-the-art methods on MultiNews, Qasper, L-Eval and RULER. For example, on the LLaMA-3.1-70B model, TTKV achieves a ROUGE-L score of 29.3, outperforming all state-of-the-art methods. This demonstrates TTKV's ability to preserve model accuracy while significantly reducing traffic and latency.

\subsection{Ablation Study}
\label{subsec:ablation}

We conduct an ablation study to evaluate the individual contributions of the fast/slow‑tier, differential quantization, streaming attention, and block size. Experiments are performed on LLaMA‑3.1‑8B with a 64K context on RULER, focusing on host‑to‑GPU traffic, p95 latency, and accuracy.

\paragraph{Fast-tier and Slow-tier (Tier Layout):}  
We compare TTKV with both the fast-tier and slow-tier to a version that stores all KV entries in a single-tier (slow-tier only). The results in Table~\ref{tab:combined_ablation} indicate that TTKV with the tiered memory architecture significantly outperforms the single-tier version. Specifically, using a single-tier configuration results in a 10\% drop in accuracy, along with a substantial increase in both host-to-GPU traffic and p95 latency. This demonstrates the importance of the fast-tier in improving both performance and efficiency, particularly in long-context tasks.

\paragraph{Differential Quantization (Tier Content):} 
This ablation evaluates the \textbf{Tier Content} principle by comparing TTKV's differential  quantization scheme to a uniform quantization baseline applied to both keys and values. The data in Table~\ref{tab:combined_ablation} shows a clear advantage for the differential approach: it reduces p95 latency by 24\% and host-to-GPU traffic by 4.3\%, while maintaining comparable accuracy. These results demonstrate the effectiveness of our proposed differential quantization in balancing performance and efficiency.

\paragraph{Streaming Attention (Tier Interaction):} 
This ablation study isolates the contribution of the \textbf{Tier Interaction} principle by evaluating TTKV with and without streaming attention. The results are shown in Table~\ref{tab:combined_ablation}: removing streaming attention leads to significantly higher p95 latency and host-to-GPU traffic, which reveals that its ability to overlap data prefetching with attention computation is critical.

\begin{table}[t]
  \centering
  \scriptsize
  \setlength{\tabcolsep}{3.5pt}
  \renewcommand{\arraystretch}{1.2}
  \begin{tabular}{@{}lccc@{}}
    \toprule
    \textbf{Ablation Variant} & \textbf{p95 Lat.} \textbf{(ms)} $\downarrow$ & \textbf{H$\rightarrow$G} \textbf{(GB)} $\downarrow$ & \textbf{Acc.} \textbf{(\%)} $\uparrow$ \\
    \midrule
    FP16 & 2450 & 91.2 & 52.0 \\
    Single-tier (no fast-tier) & 985 & 32.0 & 41.5 \\
    \textbf{TTKV (Tier Layout)} & \textbf{590} & \textbf{8.3} & \textbf{51.8} \\
    \midrule
    FP16 & 2450 & 92.1 & 52.0 \\
    Uniform Quantization (8,8) & 796 & 13.4 & 52.0 \\
    \textbf{TTKV (Tier Content)} & \textbf{590} & \textbf{8.0} & \textbf{51.8} \\
    \midrule
    FP16 & 2450 & 92.0 & 52.0 \\
    No Streaming Attention & 1021 & 47.5 & 51.8 \\
    \textbf{TTKV (Tier Interaction)} & \textbf{590} & \textbf{8.1} & \textbf{51.8} \\
    \bottomrule
  \end{tabular}
  \caption{Ablation Results: p95 Latency, Host-to-GPU Traffic, and Accuracy (LLaMA-3.1-8B on RULER dataset)}
  \label{tab:combined_ablation}
\end{table}

\paragraph{Block Size in Slow-tier}  
We experiment with block sizes \(B_{\text{blk}} \in \{32, 64, 128, 256\}\) for KV reduction. As shown in Table~\ref{tab:block_size_impact}, 128 tokens offer the best latency/accuracy trade-off. However, smaller sizes improve accuracy but increase latency, and larger sizes reduce latency but hurt accuracy.

\begin{table}[t]
  \centering
  \scriptsize
  \setlength{\tabcolsep}{4pt}
  \renewcommand{\arraystretch}{1.2}
  \begin{tabular}{@{}rccc@{}}
    \toprule
    \textbf{Block Size} & \textbf{p95 Lat.}  \textbf{(ms)} $\downarrow$ & \textbf{H$\rightarrow$G} \textbf{(GB)} $\downarrow$ & \textbf{Acc.} \textbf{(\%)} $\uparrow$ \\
    \midrule
    32 & 1796 & 58.8 & 52.0 \\
    64 & 1150 & 50.6 & 51.9 \\
    \textbf{128} & \textbf{590} & \textbf{8.1} & \textbf{51.8} \\
    256 & 595 & 3.2 & 48.1 \\
    \bottomrule
  \end{tabular}
  \caption{Impact of Block Size in Slow-tier on p95 Latency, Host-to-GPU Traffic, and Accuracy (LLaMA-3.1-70B on RULER dataset)}
  \label{tab:block_size_impact}
\end{table}



\section{Conclusion}
\label{sec:discussion}

This paper presents \emph{TTKV}, a KV cache management framework inspired by human memory mechanisms. Unlike existing approaches that treat all KV states as equally important over time, TTKV observes that recent context requires fast and precise access, while older context can be stored more compactly and retrieved with lower fidelity. Based on this insight, TTKV organizes KV caches into temporal tiers with heterogeneous capacity, precision, and latency, aligned with the HBM–DRAM memory hierarchy.
By jointly designing tier layout, tier content, and tier interaction, TTKV effectively integrates KV reduction with memory hierarchy awareness, significantly reducing cross-tier traffic while preserving model accuracy. Experiments on multiple LLMs and long-context benchmarks show that TTKV consistently reduces end-to-end decode latency and improves throughput over state-of-the-art approaches, enabling efficient and scalable long-context LLM inference.


\bibliographystyle{named}
\bibliography{references}


\end{document}